\documentclass[10pt,twocolumn,letterpaper]{article}
\usepackage{ijcb}
\usepackage{times}
\usepackage{epsfig}
\usepackage{graphicx}
\usepackage{amsmath}
\usepackage{amssymb}

\usepackage{url}
\usepackage{bm}

\ijcbfinalcopy 

\ifijcbfinal\fi

\makeatletter
\def\ps@IEEEtitlepagestyle{
\def\@oddfoot{\mycopyrightnotice}
\def\@evenfoot{}
}
\def\mycopyrightnotice{
{\hfill \footnotesize 978-1-7281-9186-7/20/\$31.00 \copyright 2020 IEEE\hfill}
}
\makeatother

\begin{document}

\title{Fingerprint Feature Extraction by Combining Texture, Minutiae, and Frequency Spectrum Using Multi-Task CNN}

\author{Ai Takahashi$^1$, Yoshinori Koda$^{1,2}$, Koichi Ito$^1$, and Takafumi Aoki$^1$\\
1 Graduate School of Information Sciences, Tohoku University, Japan.\\
2 Biometrics Research Laboratories, NEC Corporation, Japan.\\
{\tt\small ai@aoki.ecei.tohoku.ac.jp}
}

\maketitle
\thispagestyle{empty}

\begin{abstract}
Although most fingerprint matching methods utilize minutia points and/or texture of fingerprint images as fingerprint features, the frequency spectrum is also a useful feature since a fingerprint is composed of ridge patterns with its inherent frequency band.
We propose a novel CNN-based method for extracting fingerprint features from texture, minutiae, and frequency spectrum.
In order to extract effective texture features from local regions around the minutiae, the minutia attention module is introduced to the proposed method.
We also propose new data augmentation methods, which takes into account the characteristics of fingerprint images to increase the number of images during training since we use only a public dataset in training, which includes a few fingerprint classes.
Through a set of experiments using FVC2004 DB1 and DB2, we demonstrated that the proposed method exhibits the efficient performance on fingerprint verification compared with a commercial fingerprint matching software and the conventional method.
\end{abstract}

\let\thefootnote\relax\footnotetext{\mycopyrightnotice}

\section{Introduction}

A fingerprint has the property of high permanence and high discrimination in biometric features that have been used in many biometric authentication systems, such as user authentication for mobile devices, mobile payments, and national IDs \cite{biometrics}.
In general, fingerprint matching is based on the positions and angles of the feature points, called minutiae, and the relative positions between the feature points \cite{Fingerprint-Recognition}.

Fingerprint recognition using minutiae has two problems: (i) the recognition accuracy depends on the extraction accuracy of the minutiae, and (ii) the length of features depends on the number of minutiae.
For example, in a scratchy or dry fingerprint, the break in the ridges is extracted as some minutiae.
In wet or crushed fingerprints, the number of extracted minutiae is small since the ridges are collapsed.
In order to achieve highly accurate recognition of such a fingerprint image, it is necessary to combine multiple features that can be extracted from the fingerprint image.
In addition, the fingerprint image may be captured in a different area at the time of registration and authentication since the fingerprint image is acquired by placing the finger on the scanner.
It is difficult to apply template protection to prevent leakage of biometric features since the number of extracted minutiae is different at the time of acquisition.

In order to address the above problems, fingerprint matching methods based on deep learning have been proposed \cite{ICB-2019-Li,Engelsma-IEEE-2019}.
Li et al. \cite{ICB-2019-Li} used Fully Convolutional Network (FCN) to extract feature based on the texture of fingerprint images.
Engelsma et al. \cite{Engelsma-IEEE-2019} proposed a Convolutional Neural Network (CNN) to extract texture feature of the whole fingerprint and minutia feature of the fingerprint.
It is possible to extract fixed-length fingerprint feature using CNN that are independent of the number of minutiae.
On the other hand, fingerprint matching methods using deep learning used a large-scale private fingerprint image dataset in training, and its effectiveness and reproducibility are not always guaranteed.

In this paper, we propose a new CNN-based method for extracting fingerprint feature to achieve high accuracy fingerprint recognition with training using only a public dataset.
The proposed method employs frequency spectrum as a new feature in addition to the texture and minutiae used in other fingerprint matching methods.
We introduce the minutia attention module in order to extract effective texture feature from local regions around the minutiae.
We propose new data augmentation methods, which takes into account the characteristics of fingerprint images to increase the number of images during training.
We train our CNN model using 9,000 fingerprint images (1,000 classes) of IIIT-D Multi-sensor Optical and Latent Fingerprint (MOLF) \cite{Sankaran-IEEE-2015} and show that the proposed method is more accurate than commercial fingerprint authentication software, VeriFinger, and \cite{Engelsma-IEEE-2019} through experiments using FVC2004 \cite{FVC2004}.

The contributions of this paper are summarized below.
\begin{enumerate}
\item We propose a novel CNN architecture to extract feature from fingerprint images by combining texture, minutiae and frequency spectrum.
\item We propose a minutia attention module to pay attention to the position of the minutiae to extract global and local texture features.
\item We propose a new data augmentation method specific to fingerprint images.
\item We demonstrate that the recognition accuracy is higher than that of commercial software, VeriFinger SDK, and the conventional method by using only a public dataset in training.
\end{enumerate}

\section{Related Work}

Fingerprint recognition is classified into two methods: one is based on local features, i.e., minutiae, and the other is based on global features, i.e., texture.

A commonly used local features of a fingerprint is the minutiae, which are the end points and bifurcation points derived from fingerprint ridge patterns.
The position, angle and relative relation of minutiae are used to compute the matching score.
The minutia-based method is widely used in fingerprint recognition, for example, VeriFinger SDK \cite{VeriFinger} is available as a commercial software and NIST Biometric Image Software (NBIS) \cite{NBIS} as a free software.
Although the minutiae are highly discriminative features, there is a problem that the recognition accuracy is decreased if the minutiae are not extracted correctly.
Most of the methods using global features such as textures are based on image-based template matching \cite{Fingerprint-Recognition}.
Among them, the fingerprint matching method using frequency features of images has been proposed for matching fingerprint images from which minutiae cannot be extracted, such as dry or allergic skin \cite{Ito-IEICE-2004}.
Ito et al. \cite{Ito-IEICE-2004} proposed a method called Band-Limited Phase-Only Correlation (BLPOC), which focuses on the phase information of the image.
BLPOC has improved the accuracy of matching by utilizing the fact that the texture pattern of a fingerprint has an inherent frequency band and the energy is concentrated at that frequency band.
Ito et al. \cite{IEICE-2010-Ito} also showed that fingerprint recognition can be made more accurate by combining it with a minutia-based method.
On the other hand, there is a problem that the recognition accuracy is decreased due to the rotation and nonlinear deformation between fingerprint images since only the translation between images can be considered.

Recently, fingerprint recognition methods based on CNN have been proposed to address the above problems.
Tang et al. \cite{IJCB-2017-Tang} have proposed FingerNet, which can extract minutiae from low-quality fingerprint images with high accuracy by incorporating Gabor filter-based fingerprint image enhancement into CNNs.
Through experiments using the NIST SD27 \cite{NISTSD27}, which is a latent fingerprint image dataset, they demonstrated that FingerNet exhibited higher accuracy of minutiae extraction than the commercial software.
Nguyen et al. \cite{Nguyen-ICB-2018} proposed MinutiaeNet, which is a combination of CoarseNet to detect minutiae in the whole fingerprint image and FineNet to estimate the existence probability of minutiae.
Li et al. \cite{ICB-2019-Li} proposed a fingerprint feature extractor that does not need to be aligned and is robust to rotation and translation by training FCN on a local region centered on the minutiae.
Engelsma et al. \cite{Engelsma-IEEE-2019} proposed a multi-task learning method for texture feature extraction and minutiae detection to extract fingerprint features from textures and minutiae.
Since the fingerprint features extracted using CNN are of fixed length, encryption can be applied to protect biometric information.
On the other hand, most of CNN-based methods used a large-scale private fingerprint image dataset in training, and their effectiveness and reproducibility are not always guaranteed.

\section{Method}

The CNN-based fingerprint recognition methods proposed so far use minutiae, textures, or a combination of both \cite{IJCB-2017-Tang,Nguyen-ICB-2018,Engelsma-IEEE-2019}.
To improve recognition accuracy, we consider using frequency spectrum as a feature in addition to minutiae and texture.
Since the ridge patterns of fingerprints have their inherent frequency band \cite{Ito-IEICE-2004}, it is possible to match fingerprint images that cannot extract minutiae by focusing on the frequency band.
The proposed method consists of three CNNs to extract three features: texture, minutiae, and frequency spectrum, which are then concatenated to form the fingerprint feature.
We define a loss function for each feature and train it with multi-task learning, and introduce metric learning to improve the recognition accuracy.
In order to train efficiently with fewer images, we introduce new data augmentation methods, which are specialized for fingerprint images.

\begin{figure*}[t]
  \centering
  \includegraphics[width=0.93\linewidth]{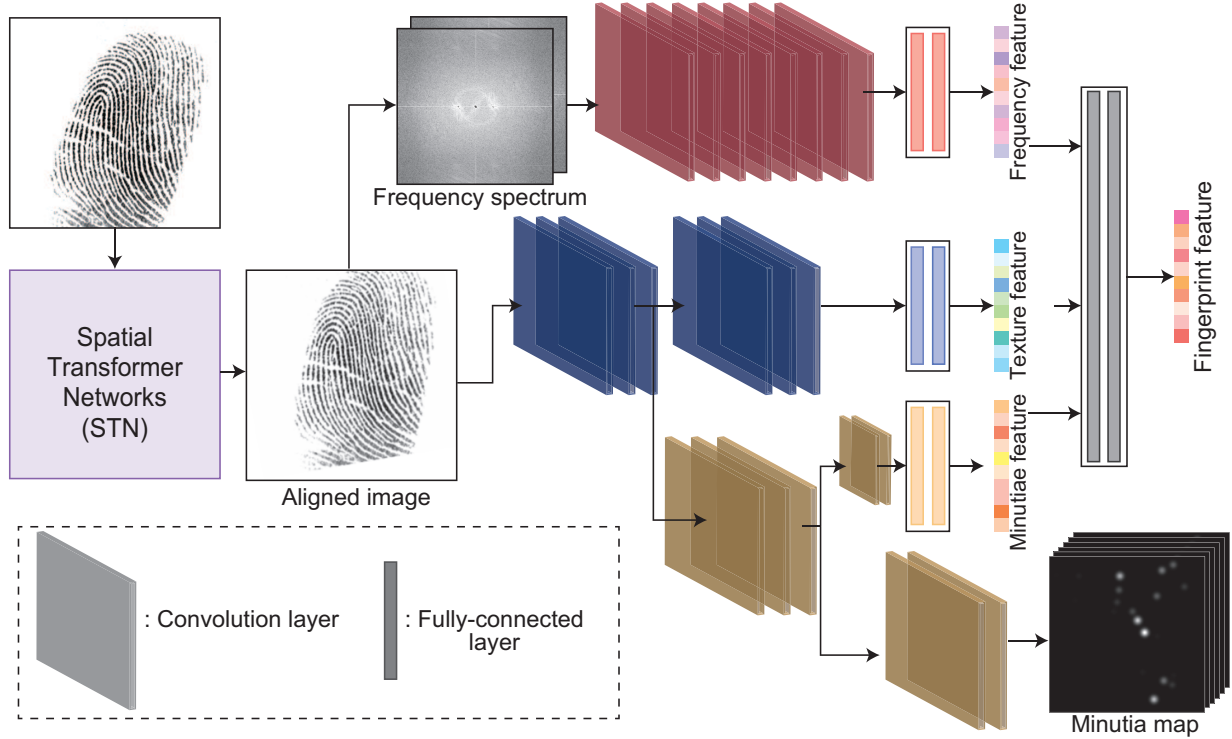}
  \caption{Overview of the network architecture used in the proposed method.}
  \label{fig:Network}
\end{figure*}

The network architecture of the proposed fingerprint feature extraction method is shown in Fig. \ref{fig:Network}.
First, we use Spatial Transformer Networks (STN) \cite{Jaderberg-NIPS-2015} to align the rotation of the fingerprint image.
Next, we extract texture-based, minutia-based, and frequency-based features from the fingerprint image using CNNs based on the Residual Network (ResNet) \cite{He-CVPR-2016}.
The three features are then concatenated to form the fingerprint feature, which is used for matching.
The details of the proposed method are described below.

\subsection{Rotation Alignment Using STN}

\begin{figure}[t]
  \centering
  \includegraphics[width=\linewidth]{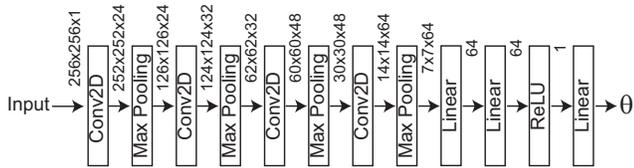}
  \caption{Architecture of localization network used in STN.}
  \label{fig:LocalizationNetwork}
\end{figure}

Before processing with the proposed method, the input fingerprint image is enhanced to increase the recognition accuracy.
In this paper, we employ the fingerprint enhancement method based on the intensity gradient, where we use the implementation of {\it FastEnhanceTexture} \cite{Enhancement}.
In the first step of the proposed method, we align the rotation of fingerprint images, which causes a decrease in accuracy in feature extraction.
In this paper, we use STN \cite{Jaderberg-NIPS-2015} for end-to-end learning of CNNs.
The STN estimates the transformation parameters in the localization network and transforms the image based on the estimated parameters.
Fig. \ref{fig:LocalizationNetwork} shows the architecture of the localization network used in the proposed method.
Although it is possible to correct the translation and rotation using the STN, we have confirmed experimentally that the correction of the rotation angle alone is sufficient.
As a result, the number of parameters to be estimated is reduced and the training is stabilized.

\subsection{Feature Extraction Using CNNs}

\begin{figure}[t]
  \centering
  \includegraphics[width=\linewidth]{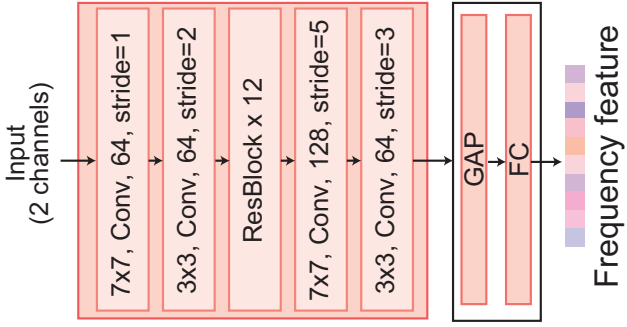}
  \caption{Architecture of CNN for extracting frequency-based feature.}
  \label{fig:FrequencyCNN}
\end{figure}

\begin{figure}[t]
  \centering
  \includegraphics[width=\linewidth]{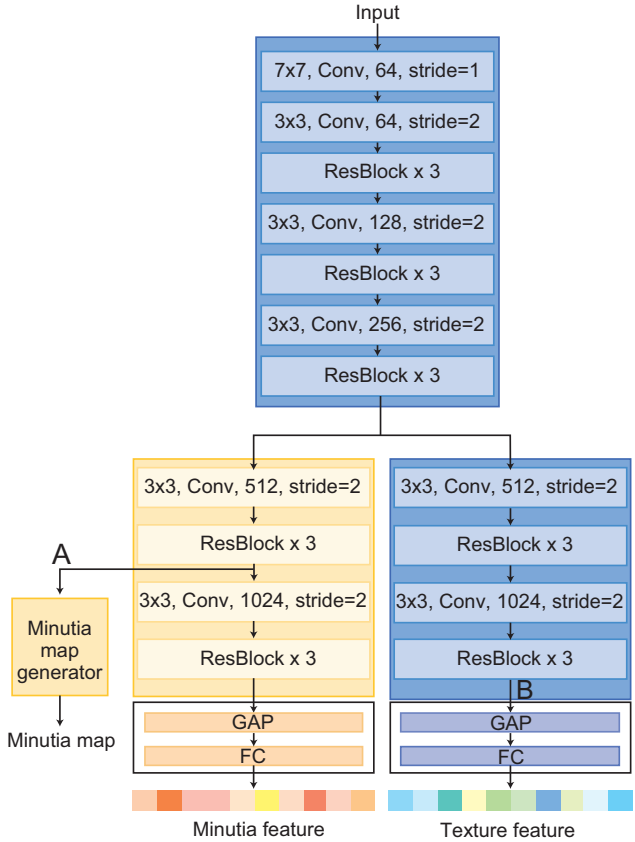}
  \caption{Architecture of CNNs for extracting minutiae-based and texture-based features.}
  \label{fig:TextureMinutiaeResNet}
\end{figure}

\begin{figure}[t]
  \centering
  \includegraphics[width=\linewidth]{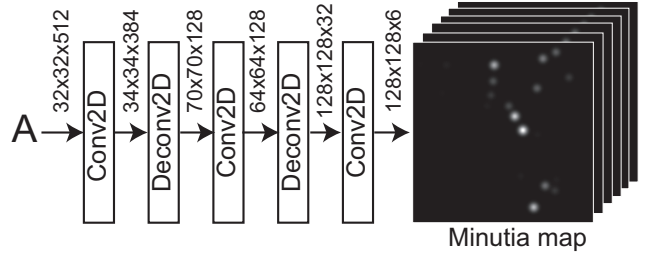}
  \caption{Architecture of minutia map generator.}
  \label{fig:MinutiaeMapCreator}
\end{figure}

Frequency, texture and minutia features are extracted from the rotation-corrected fingerprint image using ResNet \cite{He-CVPR-2016}.
The frequency features are extracted using CNN as shown in Fig. \ref{fig:FrequencyCNN}, where ResBlock is the same as the one used in ResNet \cite{He-CVPR-2016}.
The input is two channels of the real and imaginary parts of the frequency spectrum obtained by Discrete Fourier Transform (DFT) of the fingerprint image.
In addition, the following preprocessing is applied to extract the frequency features that represent the features of the fingerprint image.
The DC component is much larger than the other frequency components and represents the gain inherent to the sensor.
In order to reduce the effect of DC component, we apply normalization so that the average of the pixel values is zero, and then perform DFT.
Since the energy of the fingerprint image is concentrated in an ellipse of the low-frequency band, the high-frequency region contains only perturbations such as noise and aliasing.
In order to consider only the inherent frequency band of fingerprints similar to the BLPOC \cite{Ito-IEICE-2004}, only the region containing elliptical frequency band is extracted as an input.
The CNN network architecture for extracting texture and minutiae features is shown in Fig. \ref{fig:TextureMinutiaeResNet}.
Note that the CNNs that extract texture and minutia features share weights up to the middle.
The gray-scale fingerprint image is input to the CNN as one channel.
In order to extract minutia features using CNN, we introduce a minutia map \cite{IEEE-2019-Cao} which can represent the positions and angles of minutiae.
The minutia map is obtained with a minutiae map generator as shown in Fig. \ref{fig:MinutiaeMapCreator}, which is branched from a minutia feature extractor.


\subsection{Minutia Attention Module}

\begin{figure}[t]
  \centering
  \includegraphics[width=\linewidth]{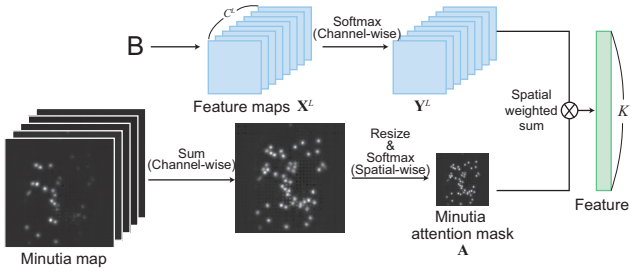}
  \caption{Architecture of minutiae attention module.}
  \label{fig:SAOL}
\end{figure}

Texture features are extracted from the entire image through training of the CNN to be effective for fingerprint recognition.
On the other hand, Li et al. \cite{ICB-2019-Li} demonstrated that using texture features in the local area around the minutiae is more accurate than using features extracted from the entire fingerprint image.
Based on this idea, we extract highly discriminative texture feature by using the location information of minutiae for texture feature extraction.
In the proposed method, we employ a Minutia Attention Module (MAM) inspired by Spatially Attentive Output Layer (SAOL) \cite{CVPR-2020-Kim} for extracting texture feature that takes into account local regions around the minutiae.
Instead of the Global Average Pooling (GAP) Layer in the texture feature extractor, MAM is used with the minutia map as an attention mask to extract features that aggregate texture information around the minutiae.
Fig. \ref{fig:SAOL} shows the architecture of MAM used in the proposed method.

Let $\bm{X}^l \in \mathbb{R}^{C^l \times H^l \times W^l}$ be the feature map output from the $l$-th layer of the texture feature extractor, where $H^l$, $W^l$, and $C^l$ are the height, width, and the number of channels, respectively, of the feature map output from the $l$-th layer ($1 \leq l \leq L$).
When using the GAP layer and Fully-Connected (FC) layer, the texture feature $\bm{t}_{\rm tex}$ is given by
\begin{equation}
    \label{eq:gap}
    \bm{t}_{\rm tex} = {\rm GAP}(\bm{X}^L)^\intercal\bm{W}^{\rm FC},
\end{equation}
where ${\rm GAP}(\bm{X}^L) \in \mathbb{R}^{C^L \times 1}$ indicates the spatially aggregated feature vector by GAP, $\bm{W}^{\rm FC} \in \mathbb{R}^{C^L \times K}$ indicates the weight matrix of the output FC layer, $(\cdot)^\intercal$ indicates the transposition, $C^L=1,024$, and $K=512$.
On the other hand, when using MAM instead of GAP, the texture feature with attention, $\bm{t}^{\rm att}_{\rm tex}$, is given by
\begin{equation}
    \label{eq:saol_1}
    \bm{t}^{\rm att}_{\rm tex}
    = {\rm MA}(\bm{X}^L)^\intercal\bm{W}^{FC},
\end{equation}
where ${\rm MA}(\bm{X}^L) \in \mathbb{R}^{C' \times K}$ is defined by
\begin{equation}
    \label{eq:saol_2}
    {\rm MA}_{c'}(\bm{X}^L) = \sum_{i,j} A_{ij} (Y_{c'})_{ij},
\end{equation}
for each $c'$, where $c'$ indicates the class index $(1 \leq c' \leq C')$, $\bm{A} \in [0,1]^{H^L \times W^L}$ indicates the attention mask, $\bm{Y} \in [0,1]^{C' \times H^L \times W^L}$ indicates the spatial logits, which is obtained by applying softmax to $\bm{X}^L$, and $(Y_c')_{ij}$ indicates the $(i,j)$-th element of $c'$-th feature map of $Y_c'$.
Note that the size of $\bm{W}^{FC}$ in Eq. (\ref{eq:saol_1}) is changed to $\mathbb{R}^{C' \times K}$ since the training is performed as 1,000-class classification problem, i.e., $C'=1,000$ in this paper.

\subsection{Loss Function and Deep Metric Learning}

This section describes the loss functions to be trained in the proposed method and deep metric learning to improve the accuracy of fingerprint recognition.

We use the MSE loss for estimating the minutia map.
Let the ground-truth and estimated minutia maps be $\bm{H}_g$ and $\bm{H}_e$, respectively.
The loss function for minutia map estimation is defined by
\begin{equation}
    \label{eq:MSELoss}
    L_{\rm map} = \sum_{i,j,k} \rho \left\{H_g (i,j,k) - H_e (i,j,k) \right\}^2,
\end{equation}
where $(i,j,k)$ indicates $(x,y)$ coordinates and $k$ channel, respectively, and $\rho$ is a constant ($\rho = 100$ in this paper).
The loss functions for texture, minutia, and frequency features are defined by the cross-entropy loss and are denoted by $L_t$, $L_m$, and $L_f$, respectively.
The loss function of the proposed method $L_{\rm all}$ is given by
\begin{equation}
    \label{eq:loss}
    L_{\rm all} = L_t + L_m + L_f + {\lambda}_{\rm map} L_{\rm map},
\end{equation}
where $\lambda_{\rm map}$ is the weight for $L_{\rm map}$.
Note that the weight parameters for STN are also optimized so as to minimize $L_{\rm all}$.

The proposed method employs deep metric learning in training to improve the accuracy of fingerprint recognition.
In this paper, we use AdaCos \cite{CVPR-2019-Zhang}, which has been demonstrated to be effective in face recognition.
The use of AdaCos makes it possible to increase the cosine similarity of the genuine pairs and to decrease that of the impostor pairs during training.
Let the class weight consisting of the center feature vector for each class be $\bm{W}$.
The probability $P_{i,y_i}$, which the $i$-th input image is classified into the class label $y_i$ of $\bm{W}$ is given by
\begin{equation}
    \label{eq:AdaCos}
    P_{i,y_i} =
    \frac{\exp \left(\tilde{s}_d^{(t)} \cdot \cos \theta_{i,y_{i}} \right)}{\exp \left(\tilde{s}_d^{(t)} \cdot \cos \theta_{i,y_{i}} \right)+\displaystyle\sum_{k \neq y_{i}} \exp \left(\tilde{s}_d^{(t)} \cdot \cos \theta_{i,k} \right)},
\end{equation}
where $\cos \theta_{i,y_i}$ indicates the cosine similarity between the $i$-th input feature and the corresponding class label $y_i$ of $\bm{W}$, $\tilde{s}_d^{(t)}$ indicates the scaling parameter, $k$ indicates the class label of $\bm{W}$ except for $i$, and $t$ indicate the number of updates.
In the case of using AdaCos, the final loss function $L$ is given by
\begin{equation}
  \label{eq:CrossEntropyLoss}
  L = - \frac {1}{N} \sum_{i=1}^{N} \log P_{i,y_{i}}.
\end{equation}

Applying AdaCos to the proposed method, $L_t$, $L_m$, and $L_f$ are calculated by Eq. (\ref{eq:CrossEntropyLoss}).

\subsection{Data Augmentation}

\begin{figure}[t]
    \centering
    \includegraphics[width=\linewidth]{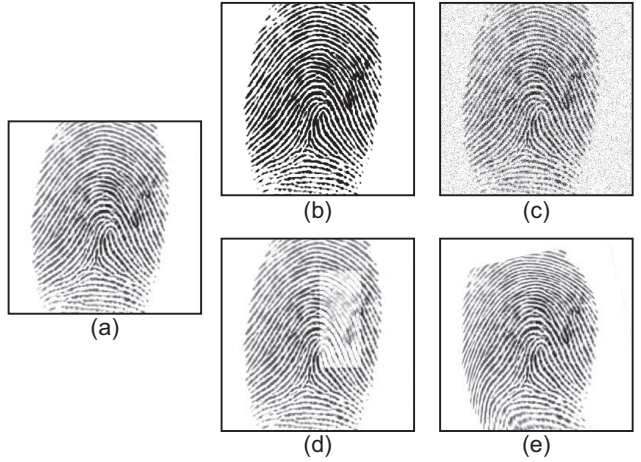}
    \caption{Example of fingerprint images after data augmentation: (a) original image, (b) random contrast, (c) random noise, (d) random morphology, and (e) random deformation.}
    \label{fig:DataAugment}
\end{figure}

In this paper, data extension is important since the training is performed only on a public dataset that is not large.
The general data augmentation methods are random contrast and random noise.
In the proposed method, we introduce random deformation and random morphology as new data augmentation methods specialized for fingerprint images.
During the acquisition process, nonlinear deformation may be added to the fingerprint image and the ridge pattern of the fingerprint may be collapsed.
In order to represent deformation of fingers, the fingerprint image is deformed according to the fingerprint deformation model \cite{Cappelli-ICAPR-2001} by random deformation.
In order to represent collapse of ridge patterns, the morphological filters of dilation and erosion are applied to a part of the fingerprint image.
Note that the size of the filter for random morphology is $0.02 \sim 0.2$\% of the image size according to random erasing \cite{Zhong-arXiv-2017}.
Fig. \ref{fig:DataAugment} shows examples of fingerprint images with data augmentation.

\subsection{Matching Score}

The matching score is calculated between fingerprint features extracted by the proposed method.
Let the fingerprint features extracted two fingerprint images be $\bm{t}_1$ and $\bm{t}_2$, respectively.
The matching score $S$ is calculated by
\begin{equation}
    \label{eq:score}
    S = \bm{t}_{1}^\intercal \cdot \bm{t}_{2}.
\end{equation}

\section{Experiments and Discussion}

This section describe the experiments of evaluating the proposed method and discuss with their results.

\subsection{Training}

The proposed CNN model is trained as follows.
We use 12,000 fingerprint images (DB1, DB2, and DB3) out of 19,200 fingerprint images in MOLF in the experiments.
Note that we do not use DB4 and DB5 consisting of latent fingerprint images since we only use the scanned fingerprint images in the experiments.
The 12,000 fingerprint images are divided into 9,000 (1,000 fingers $\times$ 3 images $\times$ 3 DBs) for training and 3,000 (1,000 fingers $\times$ 1 image $\times$ 3 DBs) for validation.
We perform multi-task learning for identifying 1,000-class labels and detecting minutiae.
RMSProp is used as an optimizer and the weight decay for preventing overfitting is set to $10^{-5}$.
The learning rates for feature extractors and STN are set to $10^{-3}$ and $5^{-4}$, respectively.
Fingerprint images are resized to $256 \times 256$ pixels as an input.
The size of the minutia map is $128 \times 128$ pixels with 6 channels.
The effective frequency band is set to 50\% of the size of frequency spectrum.
The weight for the minutia map is set to ${\lambda}_{\rm map}=10$.
The probability of random noise, random contrast, random deformation, and random morphology is set to 80\%, 80\%, 50\%, and 50\%, respectively.
We use the minutia information detected by VeriFinger SDK 10.0 \cite{VeriFinger} as the ground truth of the training data.

\subsection{Experimental Condition}

\begin{figure}[t]
    \centering
    \includegraphics[width=\linewidth]{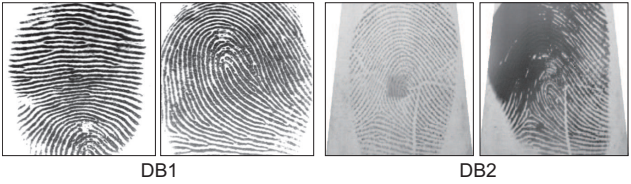}
    \caption{Example of fingerprint images from the genuine pairs in FVC2004 DB1 and DB2.}
    \label{fig:FVC2004Datset}
\end{figure}

The performance of the proposed method is evaluated using the public fingerprint image dataset, FVC2004 \cite{FVC2004} DB1 and DB2.
DB1 and DB2 consist of 800 fingerprint images (100 fingers $\times$ 8 images).
Fig. \ref{fig:FVC2004Datset} shows the example of fingerprint images from the genuine pairs in FVC2004 DB1 and DB2.
DB1 has many low-quality fingerprint images that contain little overlap in the fingerprint region, nonlinear deformation in the same subjects, collapsed ridges due to wetness, or interrupted ridges due to drying.
The DB2 has a lot of collapsed ridges, which makes it difficult to detect minutiae in many fingerprint images.
We evaluate the matching scores of 2,800 genuine pairs and 368,500 impostor pairs, which are all the combinations of the dataset.
The verification accuracy is evaluated by Equal Error Rate (EER), where is defined as the error rate where False Acceptance Rate (FAR) and False Rejection Rate (FRR) are equal.

\begin{table*}[t]
    \caption{Summary of EERs [\%] for FVC2004 DB1 and DB2.}
    \label{tb:Ablation}
    \centering
    \begin{tabular}{ccccccc|cc}
    \hline
    Method & Texture & Minutiae & Frequency & AdaCos & Data Aug. & MAM & DB1 & DB2\\
    \hline
    VeriFinger \cite{VeriFinger} &  &  &  &  &  &  & 1.63 & 1.29\\
    \hline
    Engelsma \cite{Engelsma-IEEE-2019} & \checkmark & \checkmark &  &  &  &  & 1.75 & 2.44\\
    \hline
    Proposed (A) & \checkmark & \checkmark & \checkmark &  &  &  & 1.84 & 1.84\\
    Proposed (B) & \checkmark & \checkmark & \checkmark & \checkmark &  &  & 1.69 & 1.73\\
    Proposed (C) & \checkmark & \checkmark & \checkmark & \checkmark & \checkmark &  & 1.58 & {\bf 1.10}\\
    Proposed (D) & \checkmark & \checkmark & \checkmark & \checkmark & \checkmark & \checkmark & {\bf 1.41} & 1.36\\
    \hline
    \end{tabular}
\end{table*}

\subsection{Ablation Study}

The effectiveness of refinement techniques introduced in the proposed method, i.e., frequency feature, deep metric learning, data augmentation, and MAM, is evaluated by the following ablation study using FVC2004 DB1 and DB2.
In this experiment, we compare the verification accuracy of the proposed method with Verifinger SDK 10.0 \cite{VeriFinger} and the deep learning-based method \cite{Engelsma-IEEE-2019} to demonstrate the effectiveness of the proposed method.
Note that we replaced Inception modules used in \cite{Engelsma-IEEE-2019} to ResNet modules used in the proposed method since the number of classes in the training data used in this experiment is quite small compared with the original implementation of \cite{Engelsma-IEEE-2019}, where 382,914 classes are included in the private training data.
Table \ref{tb:Ablation} shows the summary of experimental results.
In FVC2004 DB1, the EERs of the proposed method have been improved by adding refinement techniques.
The EER of the proposed method (D), which employs all the refinement techniques, is lower than that of VeriFinger and Engelsma \cite{Engelsma-IEEE-2019}.
In DB2, the accuracy of the proposed method has been also improved by adding refinement techniques.
The proposed method (C) is more accurate than VeriFinger and \cite{Engelsma-IEEE-2019}.
On the other hand, the proposed method (D) with MAM had more errors than the proposed method (C) without MAM.
This is because the fingerprint image in FVC2004 DB2 has many areas with collapsed ridges and blurred minutiae as shown in Fig. \ref{fig:FVC2004Datset}, and therefore, it may not be possible to apply proper attention to them.

\subsection{Neonate Fingerprint Recognition}

\begin{figure}[t]
    \centering
    \includegraphics[width=\linewidth]{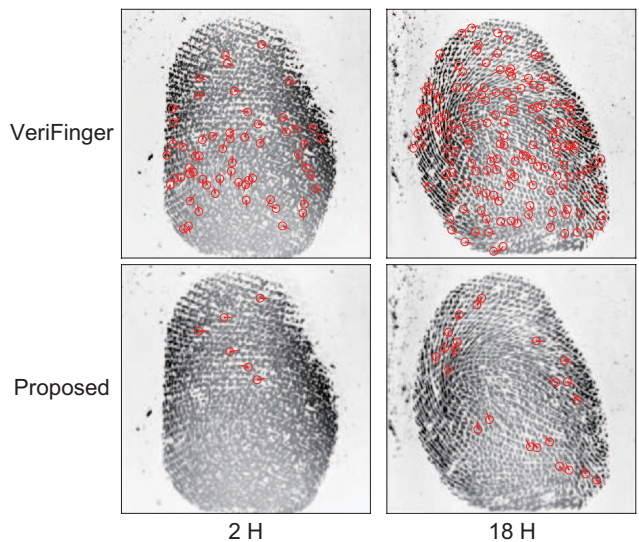}
    \caption{Minutiae extracted from neonate fingerprint images at 2 and 18 hours of age using VeriFinger SDK and the proposed method (D).}
    \label{fig:NeonateMinutiae}
\end{figure}

We evaluate the accuracy of the proposed method in neonate fingerprint recognition, which is an example of fingerprint images that are extremely difficult to extract minutiae from.
In this experiment, we use the neonate fingerprint image dataset used in \cite{BIOSIG-2019-Koda}.
This dataset contains neonate fingerprint images taken with a 1,920-ppi sensor at 2, 6, and 18 hours of age.
The number of genuine and impostor pairs is 664 and 9,632, respectively.
The EERs of VeriFinger, the proposed method (C), and (D) are 42.6\%, 34.9\%, and 38.4\%, respectively.
In VeriFinger, wrinkles and pores are extracted as minutiae as shown in Fig. \ref{fig:NeonateMinutiae}.
The EER of VeriFinger is high due to the use of many wrong minutiae for fingerprint matching.
On the other hand, the proposed method is more accurate than VeriFinger SDK in the neonate fingerprint recognition since the proposed method uses texture and frequency features in addition to minutia feature for matching.
Minutiae extracted from neonate fingerprint images using the proposed method are shown in Fig. \ref{fig:NeonateMinutiae}.
This figure shows that the proposed method was able to extract relatively correct minutiae, however, the accuracy of the proposed method (C) with MAM was reduced due to the small number of minutiae.

\subsection{Qualitative Analysis}

\begin{figure}[t]
  \centering
  \includegraphics[width=\linewidth]{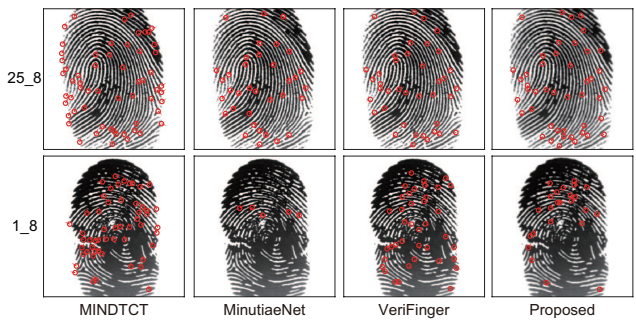}
  \caption{Example of minutia extraction from fingerprint images of FVC2004 DB1 (Upper: Quality 96, Lower: Quality 45).}
  \label{fig:MinutiaeExtraction}
\end{figure}

We present qualitative evaluation of the minutia extraction using the proposed method.
Fig. \ref{fig:MinutiaeExtraction} shows the minutiae extracted using MINDTCT \cite{NBIS}, MinutiaeNet \cite{Nguyen-ICB-2018} , VeriFinger, and the proposed method (D) for fingerprint images with quality values of 96 and 45 in VeriFinger.
For the fingerprint image with a quality value of 96, the same minutiae were extracted by all the methods.
In the case of the fingerprint image with a quality value of 45, there are many false positives for MINDTCT, and there are many undetected for MinutiaeNet, while the proposed method (D) is able to extract the majority of minutiae extracted by VeriFinger.

\begin{figure}[t]
    \centering
    \includegraphics[width=\linewidth]{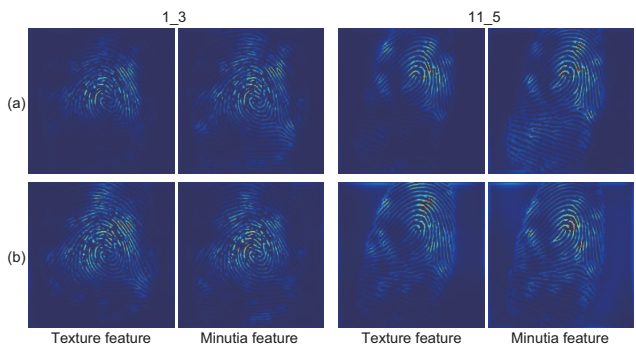}
    \caption{Saliency maps of texture and minutia features for images in FVC2004 DB1: (a) the proposed method (b) the proposed method (D).}
    \label{fig:TextureMinutiaeSaliencyMap}
\end{figure}

\begin{figure}[t]
    \centering
    \includegraphics[width=\linewidth]{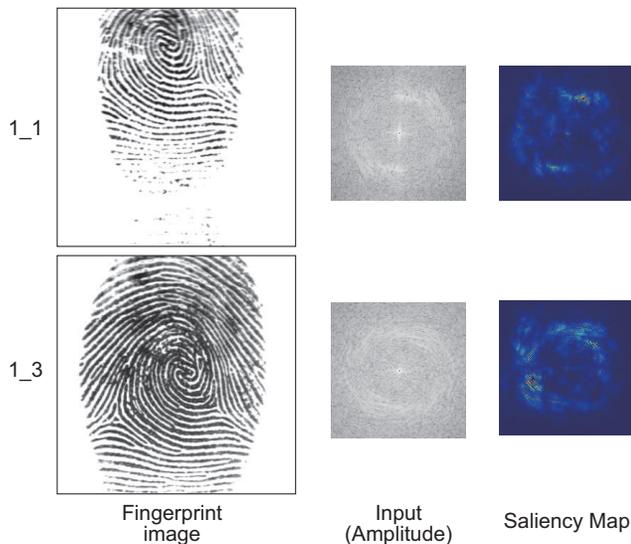}
    \caption{Saliency maps of frequency features for images in FVC2004 DB1.}
    \label{fig:FrequencySaliencyMap}
\end{figure}

The effectiveness of the proposed method is verified using a saliency map that visualizes the effect of each pixel in the input image on the extracted features.
We use Guided-back propagation \cite{Springenberg-ICLR-2015} to make a saliency map from CNN weights.
Fig. \ref{fig:TextureMinutiaeSaliencyMap} shows the saliency maps of texture and minutia features.
While CNN for extracting minutia feature focuses on the entire fingerprint image, CNN for extracting texture feature focuses on the core region.
CNN with MAM focus not only on the core region, but also on the region around minutiae.
Fig. \ref{fig:FrequencySaliencyMap} shows the saliency maps of frequency feature.
CNN for extracting frequency feature focuses on the frequency bands inherent in the fingerprint pattern and we confirmed that the individual frequency features are extracted.

\section{Conclusion}

We proposed a novel CNN architecture to extract features from fingerprint images by combining texture, minutiae and frequency spectrum.
A minutia attention module to pay attention to the position of the minutiae was proposed.
We also introduced novel data augmentation methods specific to fingerprint images for efficient training with a small number of classes of training data.
Through a set of experiments using FVC2004 DB1 and DB2, we demonstrated that the matching accuracy is higher than that of VeriFinger and the conventional method.
In future work, we will investigate the use of local frequency features and the use of relationships between local regions to recognize low quality fingerprint images from which only a few minutiae are extracted.

{\small
\bibliographystyle{ieee}
\bibliography{paperlist}

\begin{thebibliography}{10}\itemsep=-1pt

\bibitem{Enhancement}
{F}ast{E}nhance{T}exture.
\newblock
  \url{https://github.com/luannd/MinutiaeNet/blob/master/CoarseNet/MinutiaeNet%
_utils.py}.

\bibitem{FVC2004}
{FVC2004}.
\newblock \url{http://bias.csr.unibo.it/fvc2004/}.

\bibitem{NBIS}
{NIST} {B}iometric {I}mage {S}oftware ({NBIS}).
\newblock
  \url{https://www.nist.gov/services-resources/software/nist-biometric-image-s%
oftware-nbis}.

\bibitem{NISTSD27}
{NIST} {S}pecial {D}atabase 27.
\newblock
  \url{https://www.nist.gov/itl/iad/image-group/nist-special-database-2727a}.

\bibitem{VeriFinger}
{V}eri{F}inger {SDK}.
\newblock \url{https://www.neurotechnology.com/verifinger.html}.

\bibitem{IEEE-2019-Cao}
K.~Cao, D.-L. Nguyen, C.~Tymoszek, and A.~K. Jain.
\newblock End-to-end latent fingerprint search.
\newblock {\em IEEE Trans. Information Forensics and Security}, 15:880--894,
  July 2019.

\bibitem{Cappelli-ICAPR-2001}
R.~Cappelli, D.~Maio, and D.~Maltoni.
\newblock {M}odelling plastic distortion in fingerprint images.
\newblock {\em Proc. Int'l Conf. Advances in Pattern Recognition (LNCS 2013)},
  pages 371--378, Mar. 2001.

\bibitem{Engelsma-IEEE-2019}
J.~J. Engelsma, K.~Cao, and A.~K. Jain.
\newblock Learning a fixed-length fingerprint representation.
\newblock {\em IEEE Trans. Pattern Analysis and Machine Intelligence (Early
  access)}, pages 1--16, Dec. 2019.

\bibitem{He-CVPR-2016}
K.~He, X.~Zhang, S.~Ren, and J.~Sun.
\newblock Deep residual learning for image recognition.
\newblock {\em Proc. IEEE Conf. Computer Vision and Pattern Recognition}, pages
  770--778, June 2016.

\bibitem{IEICE-2010-Ito}
K.~Ito, A.~Morita, T.~Aoki, H.~Nakajima, K.~Kobayashi, and T.~Higuchi.
\newblock Score-level fusion of phase-based and feature-based fingerprint
  matching algorithms.
\newblock {\em IEICE Trans. Fundamentals}, E93-A(3):607--616, Mar. 2010.

\bibitem{Ito-IEICE-2004}
K.~Ito, H.~Nakajima, K.~Kobayashi, Aoki., and T.~T, Higuchi.
\newblock A fingerprint matching algorithm using phase-only correlation.
\newblock {\em IEICE Trans. Fundamentals}, E87-A(3):682--691, Mar. 2004.

\bibitem{Jaderberg-NIPS-2015}
M.~Jaderberg, K.~Simonyan, A.~Zisserman, and K.~Kavukcuoglu.
\newblock Spatial transformer networks.
\newblock {\em Proc. Annual Conf. Neural Information Processing Systems}, pages
  2017--2025, Dec. 2015.

\bibitem{biometrics}
A.~K. Jain, P.~Flynn, and A.~A. Ross.
\newblock {\em Handbook of Biometrics}.
\newblock Springer, 2008.

\bibitem{CVPR-2020-Kim}
I.~Kim, W.~Beak, and S.~Kim.
\newblock Spatially attentive output layer for image classification.
\newblock {\em Proc. IEEE/CVF Conf. Computer Vision and Pattern Recognition},
  pages 9533--9542, June 2020.

\bibitem{BIOSIG-2019-Koda}
Y.~Koda, A.~Takahashi, K.~Ito, A.~Takafumi, S.~Kaneko, and S.~M. Nzou.
\newblock Development of 2,400ppi fingerprint sensor for capturing neonate
  fingerprint within 24 hours after birth.
\newblock {\em Int'l Conf. Biometrics Special Interest Group (Lecture Notes in
  Informatics 296)}, pages 95--106, Sept. 2019.

\bibitem{ICB-2019-Li}
R.~Li, D.~Song, Y.~Liu, and J.~Feng.
\newblock Learning global fingerprint features by training a fully
  convolutional network with local patches.
\newblock {\em Proc. Int'l Conf. Biometrics}, pages 1--8, June 2019.

\bibitem{Fingerprint-Recognition}
D.~Maltoni, D.~Maio, A.~K. Jain, and S.~Prabhakar.
\newblock {\em Handbook of Fingerprint Recognition}.
\newblock Springer, 2003.

\bibitem{Nguyen-ICB-2018}
D.-L. Nguyen, K.~Cao, and A.~K. Jain.
\newblock {R}obust minutiae extractor: {I}ntegrating deep networks and
  fingerprint domain knowledge.
\newblock {\em Proc. Int'l Conf. Biometrics}, pages 9--16, Feb. 2018.

\bibitem{Sankaran-IEEE-2015}
A.~Sankaran, M.~Vatsa, and R.~Singh.
\newblock Multisensor optical and latent fingerprint database.
\newblock {\em IEEE Access}, 3:653--665, Apr. 2015.

\bibitem{Springenberg-ICLR-2015}
J.-T. Springenberg, A.~Dosovitskiy, T.~Brox, and M.~Riedmiller.
\newblock Striving for simplicity: {T}he all convolutional net.
\newblock {\em Proc. Int'l Conf. Learning Representations (Workshop track)},
  pages 1--14, May 2015.

\bibitem{IJCB-2017-Tang}
Y.~Tang, F.~Gao, and Y.~Liu.
\newblock Finger{N}et: {A}n unified deep network for fingerprint minutiae
  extraction.
\newblock {\em Proc. Int'l Joint Conf. Biometrics}, pages 108--116, Oct. 2017.

\bibitem{CVPR-2019-Zhang}
X.~Zhang, R.~Zhao, Y.~Quao, W.~X., and H.~Li.
\newblock Ada{C}os: {A}daptively scaling cosine logits for effectively learning
  deep face representations.
\newblock {\em Proc. IEEE/CVF Conf. Computer Vision and Pattern Recognition},
  pages 10823--10832, June 2019.

\bibitem{Zhong-arXiv-2017}
Z.~Zhong, L.~Zheng, G.~Kang, S.~Li, and Y.~Yang.
\newblock Random erasing data augmentation.
\newblock {\em CoRR}, abs/1708.04896:1--10, 2017.

\end{thebibliography}
}

\end{document}